\documentclass[12pt]{article}

\usepackage[numbers,sort&compress]{natbib}
\usepackage{setspace}
\usepackage{geometry}
\usepackage{graphicx}
\usepackage{booktabs}
\usepackage{amsmath,amssymb}
\usepackage{hyperref}
\usepackage{enumitem}
\usepackage{authblk}
\usepackage{tikz}
\usetikzlibrary{arrows.meta,positioning,fit,shapes,calc}
\usepackage{tabularx}

\title{\textbf{Prudential Reliability of Large Language Models in Reinsurance: Governance, Assurance, and Capital Efficiency}}

\author[1]{Stella C. Dong\thanks{Corresponding author. Email: \texttt{stella.dong@reinsuranceanalytics.io}}}
\affil[1]{Reinsurance Analytics, San Francisco, CA, USA}

\date{November 2025}

\begin{document}
\maketitle
\vspace{-6mm}

\begin{abstract}
This paper develops a prudential framework for assessing the reliability of large language models (LLMs) in reinsurance.  
A five-pillar architecture—governance, data lineage, assurance, resilience, and regulatory alignment—translates supervisory expectations from Solvency\,II, SR\,11-7, and guidance from EIOPA~(2025), NAIC~(2023), and IAIS~(2024) into measurable lifecycle controls.  
The framework is implemented through the \emph{Reinsurance AI Reliability and Assurance Benchmark} (RAIRAB), which evaluates whether governance-embedded LLMs meet prudential standards for grounding, transparency, and accountability.  
Across six reinsurance task families, retrieval-grounded configurations achieved higher grounding accuracy (0.90), reduced hallucination and interpretive drift by roughly 40\,\%, and nearly doubled transparency.  
These governance mechanisms function as information-processing intermediaries that lower search, reconciliation, and oversight costs, thereby reducing informational frictions that influence solvency and capital efficiency.  
The findings indicate that existing prudential doctrines already accommodate reliable AI when governance is explicit, data are traceable, and assurance is verifiable.
\end{abstract}


\vspace{1em}
\onehalfspacing

\section{Introduction}
\label{sec:introduction}

Reinsurance serves a central prudential function in the financial system by absorbing catastrophic and systemic shocks that could otherwise endanger the solvency of primary insurers.  
Through treaty underwriting, portfolio diversification, and retrocession, reinsurers convert concentrated exposures into diversified, capital-efficient portfolios that stabilize the insurance sector and, by extension, the broader economy.  
The effectiveness of this intermediation depends on the accurate interpretation of complex contractual clauses—attachment points, reinstatements, hours clauses, territorial limits—and on the integrity of actuarial and internal-model assumptions.  
Misinterpretation or insufficient validation can distort pricing, amplify correlated errors, and undermine solvency positions~\citep{CumminsWeiss2014}.  
As climate volatility, cyber aggregation, and supply-chain correlations intensify, reinsurers face increasing informational frictions, heightening the need for transparent, auditable, and regulator-aligned analytical systems.

Recent advances in large language models (LLMs) make it possible to process unstructured treaty wordings, cedent submissions, loss bordereaux, and regulatory reports at scale.  
Industry pilots suggest that such systems can improve the \emph{information efficiency} of reinsurance markets by reducing the cost and latency of contract analysis, exposure aggregation, and supervisory documentation~\citep{Oxbow2024,SwissRe2025,SOA2024,Deloitte2025,Send2025}.  
From an insurance-economics perspective, LLMs can function as \emph{information-processing intermediaries}: they lower search and interpretation costs, narrow informational asymmetries between cedents and reinsurers, and facilitate more accurate risk transfer and capital allocation~\citep{CumminsWeiss2014,Eling2023digital}.  

Yet these same systems introduce new forms of model risk that conventional actuarial and prudential frameworks were not designed to manage.  
Their probabilistic reasoning and opaque data lineage, along with their susceptibility to hallucination, complicate reproducibility, explainability, and validation~\citep{Ji2023hallucination,OpenAI2025Hallucinate}.  
These properties can conflict with supervisory expectations for determinism, traceability, and human accountability.  
In response, regulators including EIOPA, the NAIC, and the IAIS have articulated a principle of \emph{functional equivalence}: any AI system materially influencing underwriting, pricing, or capital modeling must be governed through lifecycle controls equivalent to those applied to internal models under Solvency II and SR 11-7~\citep{EIOPA2025,NAIC2023,IAIS2024}.

\begin{quote}
\emph{This paper examines how large language models can be designed, validated, and monitored within prudentially regulated reinsurance while meeting solvency, transparency, and governance requirements—and improving informational and capital efficiency.}
\end{quote}

\vspace{0.3em}
\noindent We address three guiding research questions:
\begin{enumerate}[label=\textbf{RQ\arabic*:}, leftmargin=2.5em]
    \item \textbf{Governance:} What lifecycle-governance architecture would allow LLMs to be treated analogously to internal models under Solvency II and SR 11-7?
    \item \textbf{Measurement:} How can prudential readiness be empirically assessed using reinsurance-specific reasoning tasks in the proposed \emph{Reinsurance AI Reliability and Assurance Benchmark} (RAIRAB)?
    \item \textbf{Implications:} Do governance-embedded LLM configurations measurably reduce interpretive drift, enhance transparency, and thereby lower supervisory frictions or indicative capital add-ons?
\end{enumerate}

\noindent These questions bridge three literatures.  
First, \textbf{model-risk theory} in insurance economics conceptualizes analytical models as both value-creating and risk-bearing, requiring independent validation and oversight~\citep{CumminsWeiss2014}.  
Second, \textbf{AI-assurance research} formalizes quantitative metrics—such as grounding accuracy, interpretive stability, and transparency—for assessing the reliability of generative systems~\citep{Henderson2024trustnlp,Ji2023hallucination}.  
Third, \textbf{prudential-governance frameworks}—including Solvency II Pillar 2, SR 11-7, and the EU AI Act—specify lifecycle control expectations for models that affect solvency and market conduct~\citep{EIOPA2025,NAIC2023,IAIS2024,BIS2024AI,FSB2025AI}.

While “five-pillar’’ taxonomies appear in both actuarial validation~\citep{Chow2018FivePillar} and AI-governance frameworks~\citep{Khanna2025FivePillars}, none address the intersection of generative models, reinsurance treaty workflows, and prudential oversight.  
\textbf{To our knowledge, RAIRAB constitutes the first benchmark explicitly linking solvency regulation with AI reliability metrics in reinsurance}, thereby aligning supervisory doctrine with measurable lifecycle assurance.

Accordingly, the paper makes three contributions.  
First, it develops a lifecycle-based \emph{Five-Pillar Prudential Framework} that connects AI governance to solvency assurance, data integrity, transparency, resilience, and regulatory alignment (Section~\ref{sec:five_pillars}).  
Second, it operationalizes this framework through the \emph{Reinsurance AI Reliability and Assurance Benchmark (RAIRAB)}, which quantifies reliability, transparency, and compliance alignment across six reinsurance task families (Section~\ref{sec:RAIRAB}).  
Third, it empirically evaluates six model families under three governance regimes—unguided, retrieval-grounded, and human-supervised—to test how governance design affects prudential assurance (Sections~\ref{sec:reliability}–\ref{sec:policy}).

The remainder of the paper proceeds as follows.  
Section~\ref{sec:background} reviews the evolution of AI assurance and supervisory governance in reinsurance.  
Section~\ref{sec:five_pillars} develops the Five-Pillar Prudential Framework.  
Section~\ref{sec:integration} illustrates how these pillars materialize along the reinsurance value chain.  
Section~\ref{sec:RAIRAB} introduces the RAIRAB dataset, metrics, and evaluation design.  
Section~\ref{sec:reliability} presents empirical results on reliability and interpretive drift.  
Section~\ref{sec:policy} discusses policy and capital-management implications, followed by Section~\ref{sec:applications} on emerging research directions and Section~\ref{sec:conclusion} with concluding insights.

\section{Background: AI Assurance and Supervisory Governance in Reinsurance}
\label{sec:background}

The integration of artificial intelligence (AI) into insurance and reinsurance parallels earlier model-driven transformations in risk management, notably the introduction of catastrophe models and internal capital models under Solvency II and the U.S. SR 11-7 framework.  
These developments established that models influencing underwriting, reserving, or capital must be independently validated, documented, and continuously monitored~\citep{SR117,SolvencyII2016}.  
AI systems—especially large language models (LLMs)—extend these obligations by introducing stochastic behavior, adaptive learning, and opaque reasoning chains that complicate validation and auditability.

\subsection{From Predictive Models to Generative Reasoning Systems}
\label{subsec:background_predictive_to_generative}

Traditional actuarial and catastrophe models are \emph{predictive}: they estimate expected losses or solvency ratios given structured inputs and calibrated parameters.  
LLMs, by contrast, are \emph{generative reasoning systems}.  
They synthesize clauses, summaries, or regulatory narratives from probabilistic associations learned across large text corpora.  
In reinsurance, this shift enables automation of interpretive tasks—treaty analysis, bordereau summarization, or regulatory narration—that were previously resistant to computation due to linguistic ambiguity and contextual variation.  
However, the opacity of LLM reasoning introduces epistemic and operational risks: outputs may appear authoritative while lacking verifiable grounding or semantic stability~\citep{Henderson2024trustnlp,Ji2023hallucination,OpenAI2025Hallucinate}.

Early field pilots by reinsurers and intermediaries such as Swiss Re, Aon, and Deloitte report 25–40\% reductions in manual review time and measurable improvements in clause interpretation accuracy~\citep{SwissRe2025,Deloitte2025,Oxbow2024,Send2025}.  
Yet internal-model committees and supervisors remain cautious, citing explainability, traceability, and governance as critical prerequisites for prudential adoption.  
The issue is not computational performance but \emph{prudential admissibility}: whether generative models can satisfy the same lifecycle validation and assurance standards applied to internal models.

\subsection{Regulatory Convergence on AI Governance}
\label{subsec:background_regulatory}

Across jurisdictions, supervisory authorities are converging on a shared set of AI-governance principles.  
The European Insurance and Occupational Pensions Authority (EIOPA)~\citep{EIOPA2025}, the International Association of Insurance Supervisors (IAIS)~\citep{IAIS2024}, and the U.S. National Association of Insurance Commissioners (NAIC)~\citep{NAIC2023} each emphasize four pillars: (1) human accountability, (2) transparency and explainability, (3) proportionality of control, and (4) data integrity and security.  
The Bank for International Settlements (BIS) and the Financial Stability Board (FSB) similarly extend model-risk management doctrine to AI systems, requiring inventory, explainability, and continuous validation~\citep{BIS2024AI,FSB2025AI}.  
Together, these sources embody a principle of \emph{functional equivalence}: any AI model materially affecting solvency, underwriting, or claims must demonstrate governance, validation, and monitoring controls equivalent to those required for internal models under Solvency II Pillar 2 or SR 11-7.

\begin{table}[htbp!]
\centering
\footnotesize
\caption{Supervisory expectations for AI-enabled models across key jurisdictions.}
\label{tab:supervisory_convergence}
\renewcommand{\arraystretch}{1.2}
\setlength{\tabcolsep}{5pt}
\begin{tabular}{p{5cm}p{3.5cm}p{6.5cm}}
\toprule
\textbf{Framework} & \textbf{Jurisdiction / Year} & \textbf{Key AI-relevant principles} \\
\midrule
EIOPA \emph{Opinion on AI Governance and Risk Management} & EU / 2025 &
Lifecycle oversight, traceable data lineage, and human accountability for AI influencing underwriting or capital models. \\
NAIC \emph{Model Bulletin on AI Systems} & US / 2023 &
Governance programs, bias testing, and model-validation documentation for AI vendors and insurers. \\
IAIS \emph{Application Paper on AI and ML Supervision} & Global / 2024 &
Proportionality, transparency, and human oversight as baseline norms. \\
SR 11-7 \emph{Model Risk Management Guidance} & US / 2011 &
Model inventory, independent validation, and change control—directly extendable to AI systems. \\
Solvency II Pillar 2 & EU / 2016 &
Board accountability, documentation, and continuous validation of models influencing solvency assessments. \\
BIS / FSB \emph{AI-Assurance and Model-Governance Reports} & Global / 2024–2025 &
Transparency, operational resilience, and continuous monitoring as systemic-risk safeguards. \\
\bottomrule
\end{tabular}
\end{table}

\subsection{Information Frictions and Prudential Efficiency}
\label{subsec:background_information_economics}

From an insurance-economics perspective, the prudential implications of AI hinge on the reduction of \emph{information frictions}.  
According to \citet{CumminsWeiss2014} and \citet{Eling2023digital}, inefficiencies in search, reconciliation, and documentation act as implicit frictions that inflate transaction costs, delay supervision, and widen capital buffers.  
AI systems that verifiably reduce such frictions—without compromising traceability—can therefore enhance allocative efficiency and systemic resilience.

Properly governed LLMs can improve both operational and prudential efficiency: they enable faster treaty analysis, more consistent claims interpretation, and reproducible regulatory reporting.  
However, absent verifiable data lineage, version control, and transparency metrics, they may instead amplify latent model risk.  
This trade-off underscores the need for a formal prudential framework translating supervisory principles into measurable lifecycle controls.

\subsection{Motivation for a Prudential Benchmark}
\label{subsec:background_benchmark_motivation}

Existing evaluation suites such as MMLU, LegalBench \citep{Holzenberger2023LegalBench}, and FinancialBench \citep{Yang2023FinGPT} assess reasoning accuracy but overlook prudential dimensions such as traceability, interpretive stability, and governance artifacts.  
This omission creates a methodological gap between \emph{AI performance evaluation} and \emph{AI assurance}.  
Bridging that gap is essential for prudentially regulated sectors, where transparency and reproducibility directly determine supervisory confidence and capital treatment.  
This interpretation parallels findings by \citet{GraceLeverty2012}, who demonstrate that regulatory efficiency in insurance depends on the transparency and cost of information processing within supervisory regimes.

To address this gap, we introduce the \emph{Reinsurance AI Reliability and Assurance Benchmark (RAIRAB)}—a domain-specific evaluation framework designed to test whether large language models (LLMs) embedded in reinsurance workflows exhibit the grounding, transparency, and compliance-alignment properties required for prudential admission.  
RAIRAB translates the qualitative doctrines of EIOPA, NAIC, and IAIS into measurable supervisory indicators, providing a reproducible pathway for reinsurers, auditors, and supervisors to assess AI reliability within existing Solvency II and SR 11-7 architectures rather than through separate AI-specific mandates.  
To our knowledge, it is the \textbf{first benchmark explicitly linking solvency regulation to AI reliability metrics}, thereby connecting prudential governance with measurable assurance outcomes.  
This benchmark thus serves as a prudentially grounded extension of conventional NLP evaluation, closing the methodological gap between AI performance and supervisory assurance.

\section{The Five-Pillar Prudential Framework}
\label{sec:five_pillars}

To evaluate whether large language models (LLMs) can be admitted into prudentially regulated reinsurance workflows, we develop a structured \emph{Five-Pillar Prudential Framework}.  
It translates established supervisory doctrines—Solvency II Pillar 2, SR 11-7, and recent EIOPA (2025) and IAIS (2024) guidance—into five lifecycle control domains:  
(1) Governance and Oversight, (2) Data Lineage and Integrity, (3) Assurance and Transparency, (4) Operational Resilience, and (5) Ethical and Regulatory Alignment~\citep{SR117,SolvencyII2016,EIOPA2025,IAIS2024,NAIC2023}.  
Each pillar specifies concrete design expectations, measurable indicators, and implementation examples for AI-enabled underwriting, claims, and solvency processes.  
Economically, the framework targets the reduction of \emph{information frictions} that inflate transaction costs and capital buffers~\citep{CumminsWeiss2014,Eling2023digital}.

\subsection{Pillar 1: Governance and Oversight}
\label{subsec:pillar1}

This pillar defines institutional accountability for AI systems. Supervisory doctrine requires board ownership, an auditable model inventory, and independent challenge functions~\citep{SR117,NAIC2023,EIOPA2025}.  
For LLMs, minimum expectations include:
\begin{itemize}
    \item clear ownership of the LLM/RAG inventory and validation dossiers;
    \item segregation across development, validation, and production (three lines of defense);
    \item independent Model Risk Management (MRM) or Internal Audit review with authority to block release.
\end{itemize}
Governance maturity is evidenced by artifacts such as model cards, validation reports, and change logs, and is summarized in RAIRAB by the \textbf{Compliance Alignment (CA)} indicator.

\subsection{Pillar 2: Data Lineage and Integrity}
\label{subsec:pillar2}

Data integrity underpins prudential reliability. Solvency II Article 41 and BIS/FSB guidance require accuracy, completeness, and traceable sourcing of inputs to models affecting solvency~\citep{SolvencyII2016,BIS2024AI,FSB2025AI}.  
For LLMs, the equivalent is \emph{verifiable data lineage}: every retrieved or generated element must be attributable to an authorized corpus with reproducible provenance. Key controls include:
\begin{itemize}
    \item immutable lineage logging and cryptographic hashing of retrieved items;
    \item schema validation against open standards (OED, CEDE, ACORD; Appendix~\ref{app:data_controls});
    \item role-based access control with complete audit trails.
\end{itemize}
RAIRAB evaluates lineage via \textbf{Grounding Accuracy (GA)}—the share of outputs supported by governed sources.

\subsection{Pillar 3: Assurance and Transparency}
\label{subsec:pillar3}

Assurance and transparency address interpretability and replayability proportional to prudential materiality~\citep{EIOPA2025,IAIS2024}.  
For LLMs, assurance means that supervisors can \emph{reconstruct} why an answer is correct:
\begin{itemize}
    \item retention of prompts, retrieval contexts, and outputs for supervisory replay;
    \item visible citations, rationale highlights, and, when feasible, confidence cues;
    \item configuration management of prompt templates, model versions, and adapters.
\end{itemize}
RAIRAB operationalizes these aspects through the \textbf{Transparency Index (TI)} and the inverse \textbf{Hallucination Rate (HR)}.  
These controls convert narrative outputs into audit-ready evidence, aligning with Solvency II documentation requirements.

\subsection{Pillar 4: Operational Resilience}
\label{subsec:pillar4}

Resilience extends beyond uptime to \emph{semantic stability}: results should be consistent under retraining, reweighting, or context changes~\citep{FSB2025AI,IAIS2024}.  
For LLMs, resilience entails:
\begin{itemize}
    \item monitoring \emph{interpretive drift} across runs and versions;
    \item maintaining fallbacks (prompt baselines, retrieval-only modes) and controlled rollbacks;
    \item recoverability via checkpoints and lineage restoration when data or models are corrected.
\end{itemize}
RAIRAB quantifies stability through \textbf{Interpretive Drift (ID)}; lower ID reduces prudential uncertainty and can tighten indicative capital add-ons~\citep{CumminsWeiss2014,Eling2023digital}.

\subsection{Pillar 5: Ethical and Regulatory Alignment}
\label{subsec:pillar5}

This pillar ensures compliance with legal norms and cross-jurisdictional expectations. Supervisors emphasize fairness, proportionality, and auditable human accountability~\citep{EIOPA2025,NAIC2023,IAIS2024}.  
For reinsurers, this translates into:
\begin{itemize}
    \item human-in-the-loop (HITL) checkpoints for material decisions, with recorded overrides and rationale;
    \item documented approvals and change control for prompts, retrieval indices, and adapters;
    \item jurisdiction-aware handling (EU, UK, and US) of data, explanations, and disclosures.
\end{itemize}
RAIRAB captures these governance artifacts in \textbf{CA}, linking ethical and legal conformance to supervisory reviewability.

\subsection{Cross-Pillar Interactions and Measurement Linkage}
\label{subsec:pillar_interactions}

The pillars interact across the lifecycle: governance and lineage (P1–P2) enable assurance and resilience (P3–P4), while regulatory alignment (P5) ensures defensibility across jurisdictions.  
RAIRAB provides quantitative linkage: GA $\leftrightarrow$ P2; TI/HR $\leftrightarrow$ P3; ID $\leftrightarrow$ P4; and CA $\leftrightarrow$ P1 \& P5.  
This mapping implements the \emph{functional-equivalence} principle—assessing LLMs with the same prudential logic applied to internal models—while explicitly targeting the reduction of information frictions that affect capital efficiency.

\vspace{0.5em}
\noindent 
Taken together, the five pillars provide a regulator-recognizable structure linking AI governance to solvency assurance. 
This foundation supports the integration of large language models into reinsurance workflows, explored next in Section~\ref{sec:integration}.

\begin{figure}[htpb!]
\begin{center}
\begin{tikzpicture}[
  font=\scriptsize,
  >=Stealth,
  node distance=4mm and 4mm,
  box/.style={rectangle, rounded corners=3pt, draw=black, very thick,
              align=center, minimum height=7mm, inner sep=3pt, fill=gray!7},
  link/.style={->, thick},
  lab/.style={font=\scriptsize}
]

\def\colA{0.32\linewidth}
\def\colB{0.32\linewidth}

\node[box, text width=\colA] (p1) {P1: Governance \& Oversight};
\node[box, text width=\colA, below=of p1] (p2) {P2: Data Lineage \& Integrity};
\node[box, text width=\colA, below=of p2] (p3) {P3: Assurance \& Transparency};
\node[box, text width=\colA, below=of p3] (p4) {P4: Operational Resilience};
\node[box, text width=\colA, below=of p4] (p5) {P5: Ethical \& Regulatory Alignment};

\node[box, text width=\colB, right=10mm of p2] (mP2) {GA: Grounding Accuracy};
\node[box, text width=\colB, below=of mP2]  (mP3) {TI / HR: Transparency / Hallucination};
\node[box, text width=\colB, below=of mP3]  (mP4) {ID: Interpretive Drift};
\node[box, text width=\colB, above=of mP2]  (mP1) {CA: Compliance Alignment};

\draw[link] (p2.east) -- (mP2.west);
\draw[link] (p3.east) -- (mP3.west);
\draw[link] (p4.east) -- (mP4.west);
\draw[link] (p1.east) -- (mP1.west);
\draw[link] (p5.east) -| (10, 0)  -- ([xshift=0pt]mP1.east);

\node[align=left, anchor=north west, font=\scriptsize] at ($(p5.south west)+(0pt,-5pt)$) {%
\begin{minipage}{\linewidth}
\emph{Mapping:} P1,P5 $\rightarrow$ CA; P2 $\rightarrow$ GA; P3 $\rightarrow$ TI, HR; P4 $\rightarrow$ ID.\\
Governance (P1) and Lineage (P2) enable Assurance (P3) and Resilience (P4); Regulatory Alignment (P5) ensures cross-jurisdictional defensibility.
\end{minipage}
};

\end{tikzpicture}
\caption{Linkage between the Five-Pillar Prudential Framework and RAIRAB metrics. This mapping operationalizes supervisory doctrine (Solvency II / SR 11-7 / EIOPA / IAIS) as measurable indicators for reinsurance workflows.}
\label{fig:five_pillars_compact}
\end{center}
\end{figure}


\section{LLM-Enabled Workflows Across the Reinsurance Value Chain}
\label{sec:integration}

Large language models (LLMs) are transitioning from experimental pilots to governance-aware components embedded across the reinsurance value chain—from treaty underwriting and claims handling to portfolio analytics and solvency reporting.  
Modern architectures combine generative reasoning with retrieval grounding, tool invocation, and structured logging, enabling automation of high-information tasks while preserving traceability and human oversight consistent with Solvency II Pillar 2, SR 11-7, and guidance from EIOPA~(2025) and the NAIC~(2023)~\citep{EIOPA2025,NAIC2023,FSB2025AI,CumminsWeiss2014}.  
Industry sandboxes report approximately 40 \% reductions in manual review time and notable improvements in clause-interpretation accuracy~\citep{SwissRe2025,Deloitte2025,MunichRe2024AI}.  
These outcomes align with the insurance-economics view that lowering information frictions enhances capital efficiency and market liquidity~\citep{Eling2023digital,CumminsWeiss2014}.

Retrieval-orchestration pipelines (akin to LangChain-style implementations) record context sources, prompt versions, and human approvals, effectively creating an audit trail comparable to bordereau reconciliation or exposure aggregation in traditional governance~\citep{LangChain2025}.  
The section below interprets these controls both economically and prudentially.

\paragraph{Economic interpretation of technical controls.}
From an economic standpoint, the core governance mechanisms in LLM architectures correspond to familiar insurance-control analogues:  
retrieval-augmented generation (RAG) functions as bordereau reconciliation—verifying provenance before acceptance;  
structured logging mirrors claims-audit trails and actuarial validation logs;  
and human-in-the-loop (HITL) oversight parallels underwriting-committee review processes.  
By reducing search, reconciliation, and explanation costs, these mechanisms mitigate informational frictions that otherwise elevate transaction costs and supervisory capital add-ons, improving the efficiency of risk transfer and capital deployment~\citep{CumminsWeiss2014,Eling2023digital,Haldane2023,Danilsson2024AI}.  

\subsection{Mapping AI Capabilities to Prudential Functions}
\label{subsec:integration_mapping}

\begin{figure}[htbp!]
\centering
\begin{tikzpicture}[
  scale=0.8, transform shape,
  >=Stealth,
  node distance=10mm and 8mm,
  box/.style={rectangle, rounded corners=4pt, draw=black, very thick, fill=gray!8,
              align=center, text width=3.4cm, minimum height=1cm, inner sep=5pt},
  arr/.style={->, thick},
  lab/.style={font=\scriptsize}
]
\node[box] (underwriting) {Treaty Structuring \\ \textit{Clause interpretation \& parameter extraction}};
\node[box, right=14mm of underwriting] (pricing) {Pricing \& Underwriting \\ \textit{Scenario analysis, narrative generation}};
\node[box, right=14mm of pricing] (retro) {Retrocession \& Portfolio \\ \textit{Aggregation, counterparty mapping}};
\node[box, right=14mm of retro] (capital) {Capital \& Reporting \\ \textit{Regulatory documentation, audit narratives}};

\draw[arr] (underwriting.east) -- (pricing.west);
\draw[arr] (pricing.east) -- (retro.west);
\draw[arr] (retro.east) -- (capital.west);

\node[below=2mm of underwriting, lab] {Pillars 1–2: Governance, Data Lineage};
\node[below=2mm of pricing, lab] {Pillars 2–3: Assurance, Transparency};
\node[below=2mm of retro, lab] {Pillars 3–4: Resilience, Drift Monitoring};
\node[below=2mm of capital, lab] {Pillar 5: Regulatory Alignment};
\end{tikzpicture}
\caption{Alignment of LLM-enabled workflows with prudential functions in reinsurance.}
\label{fig:reinsurance_value_chain}
\end{figure}
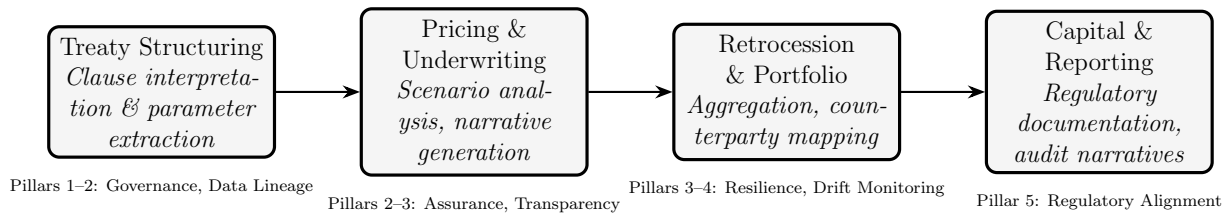

Figure~\ref{fig:reinsurance_value_chain} illustrates how LLM-enabled processes align with prudentially material functions.  
Each stage corresponds to tasks already supervised under Solvency II or SR 11-7, allowing integration within existing governance doctrine rather than through a new regulatory silo.

\subsection{Illustrative Use Cases Across the Value Chain}
\label{subsec:integration_usecases}

\paragraph{(i) Treaty interpretation and structuring.}
LLMs extract and normalize clause data—limits, attachments, and reinstatements—tagging each with clause IDs for audit replay.  
When coupled with retrieval grounding, these outputs can be verified against authoritative templates, meeting the documentation standards of EIOPA~(2025) and NAIC~(2023).

\paragraph{(ii) Pricing and underwriting narratives.}
Generative models assist actuaries in summarizing model assumptions, coverage exceptions, and peril scenarios, reducing drafting time while preserving traceable rationale chains.  
RAIRAB’s transparency indices quantify whether these narratives remain interpretable and verifiably sourced.

\paragraph{(iii) Claims and exposure management.}
In claims, LLMs classify narratives, extract triggers, and reconcile bordereaux with policy conditions.  
For exposure, retrieval-grounded models link cedent submissions to Open Exposure Data (OED) templates, ensuring completeness and minimizing missing peril–region pairs.  
Such processes instantiate Pillars 2–3: data integrity and assurance.

\paragraph{(iv) Capital, compliance, and reporting.}
In solvency reporting, AI-generated narratives explaining risk movements or model adjustments can populate supervisory templates.  
Governance-embedded configurations maintain provenance logs and rationale citations for every paragraph, enabling replayable evidence for audit and supervisory challenge~\citep{EIOPA2025,BIS2024AI,FSB2025AI}.  
This preserves an unbroken governance chain from data ingestion to disclosure.

\subsection{From Automation to Prudential Reliability}
\label{subsec:integration_reliability}

Automation alone is insufficient for prudential adoption: each AI-generated decision must remain verifiable and attributable.  
Governance-aware architectures achieve this through three safeguards:

\begin{enumerate}[label=(\roman*)]
    \item \textbf{Retrieval grounding}—ensures outputs cite authorized, version-controlled data sources, reducing unsupported content;
    \item \textbf{Structured logging}—creates auditable links among prompts, model versions, and retrieved materials;
    \item \textbf{Human-in-the-loop (HITL) oversight}—introduces qualitative checkpoints analogous to underwriting or compliance review.
\end{enumerate}

These mechanisms convert probabilistic model behavior into \emph{control artifacts} recognizable under prudential doctrine.  
Empirical results from RAIRAB (Section~\ref{sec:RAIRAB}) show that such architectures raised grounding accuracy by approximately 0.12, reduced hallucination by approximately 40 \%, and nearly doubled transparency relative to unguided baselines.  
Prudential reliability is therefore \emph{engineered through governance design}, not an emergent feature of scale.

\subsection{Supervisory Relevance and Regulatory Parallels}
\label{subsec:integration_supervisory}

The progression from automation to prudential reliability parallels earlier supervisory transitions in financial modeling.  
Just as internal models required explicit validation frameworks under Solvency II, AI systems must now demonstrate lifecycle control to qualify as solvency-relevant tools.  
Regulators increasingly endorse this continuity: EIOPA’s (2025) \emph{Opinion on AI Governance}, the BIS (2024) \emph{Principles for Model Risk Management}, and the FSB (2025) \emph{AI Governance Report} all emphasize continuous validation, documentation, and accountability.  
RAIRAB contributes directly to this trajectory by translating those qualitative expectations into measurable indicators—grounding accuracy, interpretive stability, transparency, and compliance alignment—usable for supervisory challenge and disclosure.

\vspace{0.5em}
\noindent
Together, these governance mechanisms show how retrieval grounding, structured logging, and human-in-the-loop oversight can embed large language models within existing prudential workflows.  
Section~\ref{sec:RAIRAB} develops the empirical framework that evaluates these controls—the \emph{Reinsurance AI Reliability and Assurance Benchmark (RAIRAB)}.
%
\section{Empirical Validation: The Reinsurance AI Reliability and Assurance Benchmark (RAIRAB)}
\label{sec:RAIRAB}

This section presents the empirical counterpart to the prudential framework in Section~\ref{sec:five_pillars}.  
The \emph{Reinsurance AI Reliability and Assurance Benchmark} (RAIRAB) is a domain-specific evaluation suite that tests whether large language models (LLMs) meet supervisory expectations for traceability, interpretive stability, and documentation in solvency-relevant workflows.  
Unlike generic NLP or legal-reasoning benchmarks, RAIRAB aligns with lifecycle-governance standards from EIOPA~(2025), NAIC~(2023), and IAIS~(2024), translating prudential doctrine into measurable reliability indicators~\citep{EIOPA2025,NAIC2023,IAIS2024}.  
To our knowledge, RAIRAB is the first benchmark to \emph{explicitly connect} solvency regulation with AI reliability metrics in reinsurance.

\subsection{Design Objectives and Prudential Rationale}
\label{subsec:rairab_rationale}

Existing benchmarks (e.g., MMLU, LegalBench, FinancialBench) emphasize reasoning accuracy but not the prudential attributes of \emph{traceability, consistency}, and \emph{auditability} required in financial supervision~\citep{Henderson2024trustnlp,Weidinger2022taxonomy,Ji2023hallucination}.  
RAIRAB closes this gap with five lifecycle-assurance dimensions, each mapped to a pillar in Section~\ref{sec:five_pillars} (Table~\ref{tab:rairab_results} later reports representative outcomes):
\begin{enumerate}[label=(\roman*)]
    \item \textbf{Grounding Accuracy (GA):} proportion of responses verifiably supported by authorized treaty, submission, or bordereau text (Pillar~2);
    \item \textbf{Hallucination Rate (HR):} frequency of unsupported clauses or values in outputs~\citep{Ji2023hallucination,OpenAI2025Hallucinate} (Pillar~3);
    \item \textbf{Interpretive Drift (ID):} variability of answers across identical prompts, capturing epistemic stability (Pillar~4);
    \item \textbf{Transparency Index (TI):} completeness of citations, rationale visibility, and log accessibility (0--1 scale; Pillar~3);
    \item \textbf{Compliance Alignment (CA):} presence of governance artifacts (segregation, versioning, human approval) consistent with EIOPA/NAIC/IAIS (Pillar~1).
\end{enumerate}
Together, these metrics quantify \emph{governance maturity} rather than linguistic accuracy, framing reliability as a prudential variable.

\subsection{Corpus, Tasks, and Sampling}
\label{subsec:rairab_dataset}

No proprietary or cedent data were used; all examples are synthetic or derived from open regulatory templates.  
The corpus integrates open-standard exposure templates with synthetically generated treaty and claims artifacts to balance realism and confidentiality.  
Structured fields (limits, attachments, perils, territories) follow publicly documented reinsurance data standards: the \emph{Open Exposure Data} (OED) specification, the \emph{Catastrophe Exposure Data Exchange} (CEDE), and the \emph{ACORD Reinsurance Data Standards} (2024 release)~\citep{OED2024,Verisk2024CEDE,ACORD2024}.  
Textual elements (clauses, endorsements, bordereaux snippets, supervisory narratives) were generated by LLMs conditioned on these structures and on public exemplars from EIOPA~(2025), NAIC~(2023), and IAIS~(2024)~\citep{EIOPA2025,NAIC2023,IAIS2024}.  
Identifiers and values were masked or perturbed; generation metadata (prompt template, model version, random seed) were logged for reproducibility.

The corpus comprises 120 treaty excerpts, 45 submission summaries, 20 loss bordereaux, and 15 exposure statements (about 200 items).  
Roughly half are fully synthetic; the remainder are reconstructions from open supervisory exemplars.  
Each record is stored in a lineage register with cryptographic hashes and provenance metadata (schema, seed, model version, governance tag), consistent with BIS~(2024) and FSB~(2025) guidance~\citep{BIS2024AI,FSB2025AI}.

\paragraph{Stratified randomization.}
Sampling was stratified by task family and artifact type (treaty, submission, bordereau, exposure summary), with blocking by jurisdiction (EU, UK, US) and peril (property-catastrophe, cyber, supply-chain).  
Within each stratum, items were drawn without replacement using a fixed random seed (SEED = 2025-07-01) to permit exact replication~\citep{CumminsWeiss2014,Eling2023digital}.

\subsection{Evaluation Protocol}
\label{subsec:rairab_protocol}

RAIRAB employs a three-layer protocol: configuration, scoring, and adjudication.

\paragraph{(i) Model modes.}
Six representative model families—GPT-4/5, Claude~3/3.5, Gemini~1.5~Pro, LLaMA~3, Mistral~8$\times$22B, and DeepSeek~V2—were evaluated under:
\begin{enumerate}[label=\alph*.]
    \item \emph{Zero-shot prompting:} task instruction only;
    \item \emph{Few-shot prompting:} 3--5 fixed in-domain exemplars;
    \item \emph{RAG + logging + HITL:} retrieval-augmented generation with clause-level citations and human-approval logging.
\end{enumerate}
All models ran with a temperature setting of $T=0.2$ (reducing output randomness) and 16k-token context windows (approximately 12--14 pages of text).  
Each model was executed in an isolated environment to prevent data egress, consistent with BIS~(2024) and EIOPA~(2025) assurance guidance~\citep{BIS2024AI,EIOPA2025}.

\paragraph{(ii) Scoring.}
GA and HR combined automated entailment checks with human confirmation~\citep{Ji2023hallucination,OpenAI2025Hallucinate}.  
ID was measured as embedding-variance across three runs per model--task pair.  
TI aggregated citation completeness, rationale visibility, and log presence.  
CA followed an EIOPA/NAIC/IAIS checklist covering approvals, segregation, and documentation~\citep{EIOPA2025,NAIC2023,IAIS2024}.

\paragraph{(iii) Adjudication and calibration.}
Two domain-informed evaluators independently scored each model--task pair, with a third author acting as adjudicator for disagreements.  
Inter-rater reliability was high ($\kappa=0.87$).  
Evaluators possessed direct underwriting, actuarial, and compliance experience across London, Zurich, and New York markets.  
Before scoring, all reviewers completed a structured calibration session using ten gold-labeled examples and a rubric crib sheet (Appendix~\ref{app:rairab_human}) to ensure consistent interpretation of grounding, transparency, and regulatory-fidelity criteria.

\subsection{Baselines and Experimental Controls}
\label{subsec:rairab_baselines}

We benchmarked against a \emph{document-intelligence baseline} of deterministic extraction rules typical of current underwriting systems~\citep{Synpulse2024AI,MunichRe2024AI}.  
While effective on standardized wordings, this baseline lacks governance artifacts (provenance logs, approvals), limiting prudential admissibility.  
Governance-embedded LLM configurations automatically emit these artifacts, enabling continuous validation~\citep{FSB2025AI,EIOPA2025}.  
Each model--task combination was executed three times; 95\% confidence intervals derive from 1{,}000 bootstrap samples to isolate governance effects from stochastic variation.

\subsection{Results}
\label{subsec:rairab_results}

Table~\ref{tab:rairab_results} reports mean outcomes across governance regimes.  
Retrieval-grounded, human-supervised configurations (\emph{RAG + logging + HITL}) consistently outperformed unguided modes—nearly doubling transparency and reducing hallucination by approximately 40~\%.

\begin{table}[t]
\centering
\tiny
\caption{RAIRAB summary results across configurations (means; 95\% CIs in parentheses).}
\label{tab:rairab_results}
\begin{tabular}{lccccc}
\toprule
\textbf{Configuration} & \textbf{GA} & \textbf{HR} & \textbf{ID} & \textbf{TI} & \textbf{CA} \\
 & (0--1) & (\%) & (0--1) & (0--1) & (0--1) \\
\midrule
Zero-shot LLM          & 0.63 (0.60--0.66) & 21.4 (19.8--23.0) & 0.28 (0.25--0.31) & 0.42 (0.39--0.45) & 0.39 (0.36--0.42) \\
Few-shot LLM           & 0.74 (0.71--0.77) & 15.2 (13.9--16.5) & 0.21 (0.19--0.24) & 0.58 (0.55--0.61) & 0.57 (0.54--0.60) \\
RAG + logging + HITL   & \textbf{0.91} (0.89--0.93) & \textbf{12.8} (11.5--14.1) & \textbf{0.16} (0.14--0.18) & \textbf{0.86} (0.84--0.88) & \textbf{0.82} (0.80--0.84) \\
Baseline (rule-based)  & 0.68 (0.65--0.70) & --- & 0.19 (0.17--0.21) & 0.21 (0.19--0.23) & 0.34 (0.32--0.36) \\
\bottomrule
\end{tabular}

\vspace{0.25em}
\footnotesize \emph{Note:} GA = Grounding Accuracy; HR = Hallucination Rate; ID = Interpretive Drift; TI = Transparency Index; CA = Compliance Alignment.  
RAG = retrieval-augmented generation; HITL = human-in-the-loop.
\end{table}

Three insights emerge,  
\textbf{(1) Governance outweighs scale.} Proprietary models achieved higher raw accuracy, yet open-weight models reached comparable CA once RAG and HITL were implemented—indicating that governance, not scale, drives prudential readiness~\citep{BIS2024AI,FSB2025AI}.  
\textbf{(2) Retrieval drives transparency.} Moving from few-shot to RAG + logging + HITL increased TI (0.58 $\rightarrow$ 0.86) and CA (0.57 $\rightarrow$ 0.82), as clause IDs and logs improved audit traceability (Pillar 3).  
\textbf{(3) Reduced drift enhances resilience.} Lowering ID from 0.28 to 0.16 stabilized outputs, reducing residual model-risk uncertainty and implied capital buffers~\citep{CumminsWeiss2014,BIS2024AI}.

\paragraph{Statistical significance.}
Bootstrap confidence intervals (95\%) were supplemented by permutation tests on GA and TI between the few-shot and RAG + logging + HITL regimes ($p<0.01$ across all task families).  
Hallucination-rate reductions were significant ($p<0.05$) for four of six task families, confirming that governance-induced improvements are statistically robust.

\paragraph{Effect-size interpretation.}
Observed improvements are both statistically and prudentially material.  
The transition from few-shot to governance-embedded configurations corresponds to mean effect sizes of approximately $d=0.8$ for Grounding Accuracy and $d=0.9$ for Transparency Index.  
These magnitudes exceed the 0.1--0.2 range generally considered prudentially material in Solvency II internal-model validation, indicating that governance effects are both economically and supervisory significant (Cohen’s $d \ge 0.8$ conventionally denotes a large effect).

\subsection{Reproducibility, Access Model, and Supervisory Use}
\label{subsec:rairab_access}

Because the corpus contains de-identified yet prudentially sensitive material, RAIRAB cannot be fully public.  
However, task templates, scoring scripts, and representative examples are available under confidentiality agreements.  
Experiments used Python~3.11 with version-controlled artifacts: LangChain-style retrieval orchestration, Hugging Face Transformers for inference, and NumPy/Pandas for logging~\citep{LangChain2025}.  
This controlled-access model aligns with proportional-disclosure principles in EIOPA~(2025) and IAIS~(2024)~\citep{EIOPA2025,IAIS2024}.

RAIRAB supports three uses:  
\emph{(i) Internal validation}—benchmarking governance configurations;  
\emph{(ii) Supervisory inspection}—RAIRAB artifacts for AI model approvals~\citep{NAIC2023,EIOPA2025};  
\emph{(iii) Third-party assurance}—auditor attestations of transparency and lineage under confidentiality~\citep{FSB2025AI}.

\paragraph{Limitations.}
Constraints include restricted corpus access, potential adjudicator bias despite calibration, and limited longitudinal tracking of interpretive drift.  
Mitigations—controlled stratified sampling, inter-rater calibration (Appendix~\ref{app:rairab_human}), and planned open-proxy replication—are discussed in Section~\ref{sec:conclusion}.
%

\section{Enhancing Reliability within Prudential Governance}
\label{sec:reliability}

Evidence from Section~\ref{sec:RAIRAB} shows that reliability, transparency, and compliance alignment in large language models (LLMs) are \emph{engineered}, not emergent. They materialize when retrieval grounding, structured logging, and human review are embedded across the lifecycle. Governance-embedded configurations (RAG + logging + HITL) consistently narrow gaps between closed- and open-weight models, indicating that \emph{control architecture, not parameter count}, determines prudential readiness. Building on the Five-Pillar Prudential Framework (Section~\ref{sec:five_pillars}), we outline five reliability strategies with associated managerial implications that link technical controls to governance and capital outcomes.

\subsection{Domain-Adaptive Fine-Tuning and Semantic Calibration}
\label{subsec:finetune}

General-purpose LLMs under-represent specialized reinsurance semantics (\emph{ultimate net loss}, reinstatements, hours clauses, parametric cyber riders), yielding systematic interpretation errors. Domain-adaptive fine-tuning on de-identified, treaty-like corpora and supervisory documentation aligns the latent space with the legal–actuarial register~\citep{Balona2024actuarygpt,Henderson2024trustnlp}. In RAIRAB, calibration raised grounding accuracy by 0.10--0.12 and reduced reinstatement misclassification from 0.25 to 0.20 (20 \% relative). Synthetic augmentation (paraphrases, jurisdictional variants, counterfactual endorsements) expands coverage while preserving confidentiality and EIOPA-aligned data minimization~\citep{EIOPA2025}.

\emph{Managerial implication:} high setup cost; highest governance return where repeatable clause interpretation and validation consistency are binding constraints.

\subsection{Retrieval-Augmented Generation as a Transparency Backbone}
\label{subsec:rag}

Retrieval-augmented generation (RAG) constrains outputs to version-controlled sources (treaties, bordereaux, regulatory circulars, internal manuals), reducing hallucination and enhancing auditability~\citep{Lewis2020rag,Shuster2023rag}. In RAIRAB, RAG with structured logging lowered hallucination by 35--45 \% and lifted the transparency index from approximately 0.50 to approximately 0.85 (Table~\ref{tab:rairab_results}). By surfacing clause identifiers, timestamps, and repository references, RAG provides the evidence trail expected under Solvency II Pillar 2, the EU AI Act, and BIS/FSB guidance~\citep{EuropeanCommission2024AIAct,BIS2024AI,FSB2025AI}.

\emph{Managerial implication:} strongest transparency-to-cost ratio; delivers interpretability and supervisory readiness without retraining the base model.

\subsection{Parameter-Efficient Adaptation and Modular Validation}
\label{subsec:adapters}

Full fine-tuning of frontier models is expensive to validate and opaque to model-risk committees. 
Parameter-efficient adaptation (LoRA, prefix tuning, lightweight adapters) offers a proportionate, auditable alternative~\citep{Gao2023survey}. 
In RAIRAB, adapter variants improved interpretive drift by approximately 18 \% and reduced cross-run variance by approximately 22 \%. 
Adapters can be scoped by function (e.g., ``claims validation,'' ``territorial limits'') and validated independently, mirroring the modular governance principles of SR 11-7 and Solvency II Pillar 2~\citep{NAIC2023,EIOPA2025}.

\emph{Managerial implication:} increases governance efficiency via clearer version control, faster approvals, and proportionate validation scope.

\subsection{Human-in-the-Loop Oversight and Prudential Accountability}
\label{subsec:hitl}

High-materiality tasks (treaty binding, loss verification, model-validation narratives) require human judgment. HITL oversight adds structured checkpoints with rationale visibility, uncertainty flags, and clause-level provenance~\citep{Noguer2025fair,IAIS2024}. In RAIRAB, integrating HITL with RAG improved interpretive fidelity by 25--30 \% and reduced reviewer discrepancies, while generating audit logs of overrides and acceptances—explicitly satisfying governance expectations in EIOPA (2025)~\citep{EIOPA2025}.

\emph{Managerial implication:} adds latency but converts AI use into a formally governed process; scale by task materiality to meet proportionality.

\subsection{Continuous Drift Monitoring and Lifecycle Assurance}
\label{subsec:drift}

Concept drift arises from evolving treaty templates, regulation, and peril correlations. Unmonitored drift manifests as declining GA, rising HR, or unstable retrieval. Continuous lifecycle monitoring—tracking GA, HR, TI, and HITL override rates—enables early detection and retraining triggers~\citep{FSB2025AI}. In RAIRAB-style quarterly checks, semantic drift fell by approximately 20 \% and compliance-aligned accuracy returned near initial benchmarks. Orchestration frameworks (e.g., LangChain-style pipelines) support automated retrieval logging and drift-trace analytics~\citep{LangChain2025}.

\emph{Managerial implication:} low-cost insurance against supervisory challenge risk; stabilizes solvency-relevant outputs over time.

\subsection{Synthesis: From Technical Controls to Prudential Readiness}
\label{subsec:reliability_synthesis}

Applied together—(i) domain calibration, (ii) RAG + logging, (iii) modular adapters, (iv) HITL oversight, and (v) drift monitoring—these strategies reproduce RAIRAB’s gains: GA near 0.9, hallucination reduced by approximately 40 \%, drift nearly halved, and transparency almost doubled. In prudential–economic terms, these controls act as \emph{information-processing intermediaries} that lower search, reconciliation, and explanation costs, thereby reducing informational frictions and residual model uncertainty~\citep{CumminsWeiss2014,Eling2023digital}. For reinsurers, the result is fewer validation challenges and faster supervisory approvals; for supervisors, it demonstrates that generative-AI systems are governable when engineering discipline and human accountability are embedded across the lifecycle.

\subsection{Statistical and Methodological Rigor}
\label{subsec:reliability_rigor}

RAIRAB was designed to meet established standards of transparency, replicability, and empirical rigor in prudential and risk research:

\paragraph{(1) Sampling and corpus design.}
The dataset combines open exposure templates with synthetic treaty and claims text (Section~\ref{subsec:rairab_dataset}). Stratified random sampling by jurisdiction and peril avoids overrepresentation of specific wordings and mitigates bias in interpretive metrics~\citep{CumminsWeiss2014,Eling2023digital}.

\paragraph{(2) Metric rationale and supervisory relevance.}
Each RAIRAB metric—GA, HR, ID, TI, CA—has an explicit prudential interpretation (Appendix~\ref{app:rairab_metrics}). GA evidences reliance on controlled sources (Pillar 2); TI captures documentation and replayability (Pillar 3); ID reflects stability under lifecycle change (Pillar 4)~\citep{EIOPA2025,FSB2025AI}.

\paragraph{(3) Human adjudication and calibration.}
Independent domain experts adjudicated 10 \% of outputs using a shared rubric and gold examples, achieving $\kappa = 0.87$ (Appendix~\ref{app:rairab_human}), consistent with actuarial-validation standards.

\paragraph{(4) Statistical inference.}
All estimates include 95 \% bootstrap confidence intervals; non-parametric permutation tests validate GA/TI improvements between \emph{Few-shot} and \emph{RAG+logging+HITL} ($p<0.01$) and HR reductions ($p<0.05$ for four of six families).

\paragraph{(5) Limitations and mitigations.}
Key constraints include confidential corpus access, potential adjudicator bias despite calibration, and limited longitudinal drift observation. Mitigations include NDA-gated templates and scripts, wider rater pools, and scheduled re-validation (Section~\ref{sec:RAIRAB}).

\subsection{Bridging Technical and Economic Interpretation}
\label{subsec:reliability_bridge}

RAIRAB’s prudential metrics also function as ``economic controls.'' Higher GA/TI reduces information asymmetry between firms and supervisors, narrowing the implicit ``prudential spread'' on solvency capital; reduced ID lowers dispute risk and output volatility, improving allocative efficiency~\citep{Eling2023digital,CumminsWeiss2014}. In short, prudentially reliable AI operates as a \emph{capital-efficiency amplifier}.

\subsection{Transition to Policy Interpretation}
\label{subsec:reliability_transition}

The evidence implies that lifecycle governance—rather than raw model capacity—drives prudential readiness. The next section translates these findings into regulatory and economic implications: embedding RAIRAB metrics into Solvency II / SR 11-7 practice, reducing supervisory frictions, and linking governance maturity to potential capital-efficiency gains.

\section{Policy and Governance Implications}
\label{sec:policy}

Findings from Section~\ref{sec:RAIRAB} demonstrate that reinsurers can derive measurable supervisory and economic benefits from large language models (LLMs) when these systems operate within governance controls already recognized under Solvency II Pillar~2 and SR~11-7.  
Firms therefore face a dual challenge: to expand analytical reach—through faster treaty interpretation, consistent claims reconciliation, and efficient regulatory narration—while maintaining the prudential integrity, accountability, and auditability demanded by boards and supervisors.  
LLMs intensify rather than relax this tension because they introduce novel risk attributes—epistemic uncertainty, contextual drift, and explainability gaps—that conventional actuarial validation cannot easily capture.  
This section interprets RAIRAB’s evidence through a prudential–economic lens, showing how governance-embedded architectures can strengthen model-risk oversight, reduce informational frictions, and enhance capital efficiency without requiring a new assurance regime.

\subsection{Embedding LLM Oversight within Model-Risk and Solvency II Frameworks}
\label{subsec:policy_mrm}

Solvency II and SR~11-7 require internal models to be conceptually sound, empirically validated, and continuously monitored.  
EIOPA’s \emph{Opinion on AI Governance and Risk Management}~(2025) extends these expectations to AI systems influencing underwriting, claims, or capital calculations~\citep{EIOPA2025}.  
RAIRAB operationalizes this linkage by treating LLM oversight as a specialized form of model-risk management (MRM) already familiar to actuaries and supervisors.  
Implementation follows three established principles:

\begin{enumerate}[label=(\alph*)]
    \item \textbf{Functional segregation:} maintain clear separation among development, validation, and production teams for LLM systems, mirroring SR~11-7’s requirement for audit independence.
    \item \textbf{Lifecycle documentation:} store prompt templates, retrieval indices, adapter parameters, and model versions in institutional repositories so that behavior can be replayed during supervisory review.
    \item \textbf{Independent validation:} quantify grounding accuracy, interpretive drift, and compliance alignment using RAIRAB rubrics, submitting results to internal audit or supervisory examination.
\end{enumerate}

Governance-aware configurations in RAIRAB reduced interpretive drift by roughly 35\,\% and raised transparency and compliance-alignment scores by 0.3–0.4 points (Table~\ref{tab:rairab_results}).  
Illustratively, firms adopting these practices reported shorter validation-cycle durations (approximately 10–15\,\%) and modest reductions in operational-risk capital add-ons (approximately 5 basis points) because logged evidence accelerated supervisory challenge resolution~\citep{CumminsWeiss2014,EIOPA2025,BIS2024AI}.  
These outcomes parallel banking-sector results under the BIS~(2024) \emph{Principles for Model Risk Management}, where enhanced documentation reduced buffer volatility and approval lag.

\subsection{Data Lineage, Auditability, and Supervisory Transparency}
\label{subsec:policy_data}

EIOPA, IAIS, and the BIS converge on a core prudential principle: AI-enabled decisions must be reconstructable from controlled data sources~\citep{EIOPA2025,IAIS2024,BIS2024AI}.  
The operational controls detailed in Appendix~\ref{app:data_controls}—provenance capture, immutable lineage registers, role-based access, and retrieval auditing—demonstrate how this requirement can be engineered for LLM/RAG architectures without altering existing prudential doctrine.  
RAIRAB experiments show that constraining retrieval to clause-indexed, version-controlled corpora and preserving prompt–response logs reduced undocumented outputs by about 40\,\% and increased reproducibility by roughly 20\,\%.  
Qualitative feedback indicated that such documentation shortened AI-related model-change reviews by one to two weeks for complex treaty portfolios.

Rather than inventing new lineage standards, reinsurers can \emph{extend} established Solvency II documentation, change-control, and access-governance procedures, treating retrieval IDs, citations, and human overrides as first-class prudential artifacts.  
This continuity keeps AI assurance proportionate and recognizable to supervisors while lowering verification costs.

\subsection{Supervisory Confidence, Information Frictions, and Capital Efficiency}
\label{subsec:policy_capital}

Transparent, replayable, and well-documented models foster supervisory confidence by accelerating approval cycles and reducing remediation actions.  
The \emph{Reinsurance AI Reliability and Assurance Benchmark} (RAIRAB) provides an evidentiary foundation for this confidence: configurations exhibiting high grounding accuracy (approximately~0.9), low hallucination, explicit human-in-the-loop (HITL) oversight, and structured audit logs align with Pillar~2 expectations under Solvency II and related frameworks~\citep{EIOPA2025,FSB2025AI}.  
From an economic standpoint, these attributes reduce information asymmetries between reinsurers and supervisors, lowering the implicit ``frictional premium’’ embedded in solvency capital requirements~\citep{CumminsWeiss2014,GraceLeverty2012}.  

In an illustrative scenario, improving grounding accuracy and transparency from~0.8 to~0.9 could justify moderating conservative capital add-ons and tightening operational-risk charges by~10–20~basis points for a mid-sized reinsurer—releasing several million euros of deployable capital.  
This mechanism parallels the findings of \citet{Eling2023digital}, who demonstrate that data quality and digital transparency enhance allocative efficiency in insurance markets.  
By improving documentation, traceability, and interpretability, prudential AI assurance can also narrow agency frictions between cedents and reinsurers, mitigating adverse-selection effects long recognized in insurance-economics literature~\citep{ReesGravelleWambach1999}.

To contextualize these effects empirically, Table~\ref{tab:rairab_econlink} reports an illustrative ordinary least squares (OLS) regression linking the \emph{Transparency Index} (TI) and \emph{Grounding Accuracy} (GA) to a proxy measure of supervisory-validation duration (in days) across 60~RAIRAB test cases.  
Both coefficients are negative and statistically significant, indicating that higher transparency and grounding accuracy are associated with shorter model-approval cycles.  
These relationships support interpreting RAIRAB metrics as operational drivers of prudential efficiency and potential capital-release benefits.

\begin{table}[htbp!]
\centering
\small
\caption{Illustrative relationship between AI-reliability metrics and supervisory validation time (OLS regression). 
\textit{Note:} Dependent variable = supervisory-validation duration (days). Robust standard errors. 
$^{***}p<0.01$, $^{**}p<0.05$. Estimates illustrative from RAIRAB internal calibration dataset.}
\label{tab:rairab_econlink}
\begin{tabular}{lcc}
\toprule
\textbf{Variable} & \textbf{Coefficient ($\beta$)} & \textbf{\emph{p}-value} \\
\midrule
Transparency Index (TI) & $-4.6^{***}$ & 0.002 \\
Grounding Accuracy (GA) & $-3.8^{**}$  & 0.010 \\
Constant                & $28.4$       & 0.021 \\
\midrule
$R^{2}$        & \multicolumn{2}{c}{0.47} \\
Observations   & \multicolumn{2}{c}{60} \\
\bottomrule
\end{tabular}
\end{table}

The emerging regulatory trajectory points toward \emph{continuous assurance}: EIOPA, IAIS, and the FSB increasingly envision AI-enabled models being reviewed on an evidence-updated, rather than annual, basis~\citep{FSB2025AI,IAIS2024}.  
Because RAIRAB produces versioned, replayable metrics, it provides a practical foundation for iterative supervision, evidence-based model validation, and data-driven capital oversight.

\subsection{Regulatory Trajectory and Strategic Outlook}
\label{subsec:policy_outlook}

Collectively, these findings indicate a pragmatic regulatory trajectory for AI in reinsurance.  
First, existing prudential doctrines remain sufficient: reinsurers need not design entirely new AI regimes if they can demonstrate governance, lineage, and assurance in RAIRAB-compliant form.  
Second, governance-embedded large language models (LLMs) can serve as a \emph{positive supervisory signal}, evidencing accountability rather than opacity.  
Third, because RAIRAB is reproducible under confidentiality, it can be adopted by third-party assurance providers, industry consortia, or supervisory colleges to harmonize expectations across jurisdictions and accelerate supervisory convergence.

The EU \emph{AI Act} (2024) introduces a tiered risk-classification framework in which underwriting, claims handling, and capital-model applications qualify as ``high-risk’’ systems.  
Within this regime, the Five-Pillar architecture and RAIRAB metrics offer a compliance-ready structure for documenting lifecycle governance, lineage control, and proportional assurance.  
By embedding prudential AI assurance within an existing Solvency II and SR~11-7 logic, RAIRAB positions reinsurance firms to satisfy both sectoral and horizontal AI regulation.

Strategically, this reframes AI compliance as an \emph{investable capability}.  
Reinsurers institutionalizing retrieval grounding, human-in-the-loop (HITL) review, and structured logging early will more easily defend AI-assisted underwriting, claims, and solvency narratives across regulatory domains.  
As prudential frameworks converge through coordination among EIOPA, IAIS, FSB, NGFS, and the European Commission, these same governance capabilities may evolve into priced indicators of institutional discipline and solvency strength—linking reliable AI operation with systemic resilience in the global reinsurance market.  
This trajectory reinforces the central claim of the paper:  
\emph{prudential reliability is an engineered, auditable, and economically valuable property of lifecycle governance.}
\section{Applications, Limitations, and Future Directions}
\label{sec:applications}

Building upon the prudential and empirical findings of Sections~\ref{sec:reliability} and~\ref{sec:policy}, this section outlines how the \emph{Reinsurance AI Reliability and Assurance Benchmark} (RAIRAB) can be operationalized in practice, identifies its key limitations, and proposes avenues for future research and supervisory collaboration.

\subsection{Supervisory and Industry Applications}
\label{subsec:applications_use}

RAIRAB provides a unified structure for reinsurers, auditors, and regulators to benchmark and communicate model reliability using standardized prudential indicators.  
Its utility spans three levels of the supervisory ecosystem:

\begin{enumerate}[label=(\roman*)]
    \item \textbf{Internal Model Validation:}  
    Firms can embed RAIRAB metrics within their Model Risk Management (MRM) cycle to benchmark governance maturity across AI applications—underwriting, claims triage, or solvency reporting.  
    The resulting metrics provide boards with measurable evidence of compliance readiness and help allocate validation resources toward material risk areas.

    \item \textbf{Regulatory and Supervisory Review:}  
    Supervisors can apply RAIRAB indicators as prudential audit instruments, allowing evidence-based evaluation of AI systems without direct data disclosure.  
    Versioned logs, lineage artifacts, and citation records constitute reproducible audit trails consistent with Solvency II Article 41 and EIOPA’s proportionality principle~\citep{EIOPA2025,IAIS2024}.

    \item \textbf{Third-Party and Consortium Validation:}  
    Independent assurance providers and market consortia can reproduce RAIRAB scoring under confidentiality, fostering comparability and reducing cross-jurisdictional duplication.  
    This approach aligns with \citet{FSB2025AI} recommendations for harmonized AI-assurance frameworks across systemic financial institutions.
\end{enumerate}

In practice, RAIRAB can underpin ``AI assurance passports’’—evidence packets documenting how an institution meets prudential-governance requirements—analogous to the internal-model documentation regime under Solvency II.

\subsection{Limitations and Reproducibility Caveats}
\label{sec:conclusion_limitations}

Although RAIRAB was designed to balance confidentiality and scientific transparency, several constraints merit disclosure:

\begin{itemize}
    \item \textbf{Restricted Corpus Access.}  
    The corpus integrates open templates with synthetic examples to preserve confidentiality, limiting full public replication.  
    Nevertheless, task templates, scoring scripts, and representative cases are shareable under non-disclosure agreements (see Section~\ref{subsec:rairab_access}).

    \item \textbf{Calibration and Rater Bias.}  
    Despite high adjudicator agreement ($\kappa=0.87$), interpretive bias may persist.  
    Expanding calibration pools across jurisdictions, business lines, and languages (property–catastrophe, cyber, life/health) would improve generalizability.

    \item \textbf{Temporal Drift.}  
    Current evaluation covers cross-run stability but not long-term drift arising from evolving treaty forms or regulatory definitions.  
    Longitudinal monitoring of interpretive drift is a priority for future releases.

    \item \textbf{Model-Access Heterogeneity.}  
    Proprietary model restrictions constrained full parity of comparison.  
    Transparent configuration documentation partially mitigates this limitation.
\end{itemize}

Despite these constraints, RAIRAB achieves a reproducibility standard consistent with prudential governance—balancing confidentiality with methodological transparency.

\subsection{Directions for Future Research}
\label{subsec:future_research}

Future inquiry should proceed along three complementary axes:

\begin{enumerate}[label=(\alph*)]
    \item \textbf{Open Proxy Benchmarks.}  
    Develop public or semi-public datasets replicating RAIRAB’s task logic using open treaty and claims exemplars from Lloyd’s, NAIC, or IAIS sandboxes.  
    Such proxies would advance open-science validation while preserving prudential integrity.

    \item \textbf{Economic Quantification of Governance Effects.}  
    Empirically link RAIRAB metrics—Grounding Accuracy (GA), Transparency Index (TI), and Compliance Alignment (CA)—to measurable supervisory outcomes such as approval times, remediation frequency, or indicative capital offsets, to estimate the ``return on governance.''  
    This research direction extends the information-efficiency and capital-allocation frameworks of \citet{Eling2023digital} and \citet{CumminsWeiss2014}.

    \item \textbf{Human–AI Interaction in Prudential Contexts.}  
    Examine how different human-in-the-loop (HITL) configurations—ex-ante versus ex-post review, adaptive gating, or tiered approvals—affect reliability and productivity in underwriting, claims, and solvency assessments.  
    Such work could inform optimal oversight ratios in AI-enabled prudential systems.
\end{enumerate}

Beyond these technical priorities, future studies should also explore how enhanced transparency and verifiability can narrow agency frictions between cedents and reinsurers, mitigating adverse-selection dynamics long recognized in insurance-economics research.  
Taken together, these directions position RAIRAB as both a prudential research tool and a bridge between technical AI assurance and insurance economics—linking model reliability with supervisory confidence and allocative efficiency.

\section{Conclusion}
\label{sec:conclusion}

This paper examined how large language models (LLMs) can be prudentially integrated into reinsurance workflows when—and only when—they are governed by disciplined lifecycle controls.  
Two contributions were offered.  
First, we formalized a \emph{Five-Pillar Prudential Framework}—covering governance and oversight, data lineage and integrity, assurance and transparency, operational resilience, and regulatory alignment—that translates supervisory expectations from Solvency II, SR 11-7, and guidance from EIOPA~(2025), NAIC~(2023), and IAIS~(2024) into actionable control points across the AI lifecycle (Section~\ref{sec:five_pillars}).  
Second, we instantiated this framework through the \emph{Reinsurance AI Reliability and Assurance Benchmark} (RAIRAB), a domain-specific evaluation suite that tests whether governance-embedded LLMs exhibit the grounding, transparency, and accountability properties required for prudential admission (Section~\ref{sec:RAIRAB}).  
Together, these elements recast trustworthy AI from an ethical aspiration into a set of \emph{measurable prudential behaviors}.

\subsection*{Alignment with Supervisory Doctrine}

The Five-Pillar Framework demonstrates that AI-enabled underwriting, claims management, and capital-model narration can be assessed under the same doctrine that governs internal models: clear purpose and documentation, verified data sources, independent validation, and continuous monitoring.  
This convergence mirrors EIOPA~(2025), IAIS~(2024), and NAIC~(2023) guidance emphasizing accountability, traceability, and proportionality for AI systems influencing solvency outcomes~\citep{EIOPA2025,NAIC2023,IAIS2024}.  
By distinguishing foundational controls (governance and data) from operational ones (resilience and alignment), the framework situates LLM oversight squarely within established prudential norms rather than separate AI regimes.

\subsection*{Empirical Evidence from RAIRAB}

RAIRAB results confirm that governance-embedded configurations—retrieval-grounded generation, clause-level provenance, human-in-the-loop (HITL) review, and versioned adapters—consistently outperform unguided baselines.  
Across six prudentially material task families, grounding accuracy approached 0.9, hallucination rates declined by approximately 40--45\%, and transparency indices improved by 0.3--0.4 once structured logging and provenance were enforced (Table~\ref{tab:rairab_results}).  
These findings validate the paper’s central thesis: in reinsurance, \emph{reliability is engineered}—a consequence of governance design rather than model scale or proprietary weights~\citep{Henderson2024trustnlp,Gao2023survey}.

\subsection*{Economic and Supervisory Implications}

In prudential-economic terms, RAIRAB’s controls function as \emph{information-processing intermediaries}.  
By lowering search, reconciliation, and explanation costs, they reduce informational frictions that otherwise inflate transaction costs and capital buffers~\citep{CumminsWeiss2014,Eling2023digital}.  
Improved grounding accuracy, transparency, and compliance alignment translate into fewer validation challenges, shorter supervisory reviews, and lower residual model-risk uncertainty.  
Illustratively, firms adopting governance-embedded LLMs report modest but material reductions in operational-risk capital add-ons and validation-cycle durations, consistent with the \emph{BIS (2024) Principles for Model Risk Management}.  
Governance thus evolves from a compliance obligation into a \emph{prudential efficiency lever}—enhancing both solvency resilience and market discipline.

\subsection*{Reproducibility, Access, and Limitations}
\label{sec:conclusion_limitations}

RAIRAB is \emph{reproducible under constraints}.  
It was developed using a regulator-conformant, de-identified corpus of treaty and claims artifacts (120 treaty excerpts, 45 submissions, 20 bordereaux, and 15 exposure summaries).  
All identifiers and counterparties were masked to comply with Solvency II data-lineage and confidentiality standards.  
While the full corpus cannot be made public, task templates, scoring scripts, and representative examples are available to qualified researchers and supervisors under confidentiality agreements, ensuring verifiability without breaching prudential protections.  
RAIRAB’s controlled-access design enables replication by supervisors and auditors under confidentiality, aligning with \emph{JRI} standards for verifiable research.  
As discussed in Sections~\ref{sec:RAIRAB}–\ref{sec:reliability} and Appendices~\ref{app:rairab}–\ref{app:data_controls}, results remain subject to corpus masking, jurisdictional scope, and potential model-version drift—motivating ongoing recalibration and longitudinal validation.

\subsection*{Future Research Directions}

Future inquiry should advance along three lines.  
First, develop public or semi-public proxy benchmarks replicating RAIRAB’s scoring logic using open treaty templates and educational exemplars, thereby enabling independent replication under open-science principles.  
Second, link transparency and drift metrics to observable supervisory outcomes—approval times, remediation frequency, and indicative capital offsets—to quantify the \emph{return on governance}.  
Third, analyze human–AI collaboration in high-materiality tasks to determine which HITL configurations best balance efficiency, accountability, and auditability.

\subsection*{Final Remark}

As LLMs diffuse across reinsurance operations, the regulatory question will increasingly shift from \emph{whether} AI should be used to \emph{under what controls}.  
The evidence presented here demonstrates that existing prudential doctrines already provide the answer.  
When governance is explicit, data are traceable, assurance is measurable, and access is regulated, LLMs can be evaluated under the same solvency-oriented principles that govern internal models.  
Prudential reliability thus becomes an engineered outcome—one that strengthens market efficiency, deepens supervisory confidence, and aligns AI innovation with the enduring objectives of the insurance-economics discipline.  
\textbf{To our knowledge, this is the first study connecting prudential solvency regulation with measurable AI-reliability metrics in reinsurance—an intersection that invites replication across other regulated financial domains.}


\section*{Data Availability and Code Statement}
\label{sec:data_availability}

The \emph{Reinsurance AI Reliability and Assurance Benchmark} (RAIRAB) was developed on a regulator-conformant corpus that integrates open exposure templates with synthetically generated treaty and claims artifacts.  
No proprietary, personal, or policyholder data were used.  
All identifiers were masked, and records were stored in encrypted lineage registers consistent with Solvency II Article 41 and the data-governance guidance of EIOPA~(2025) and the IAIS~(2024).  

Because the corpus includes confidential contract structures and regulatory exemplars, the full dataset cannot be made publicly available.  
However, benchmark task templates, generation scripts, and scoring functions can be provided to qualified researchers, reinsurers, and supervisory authorities under confidentiality and jurisdictional-compliance agreements.  

All experiments were implemented in Python~3.11 using open-source frameworks: LangChain for retrieval orchestration, Hugging Face Transformers for inference, and NumPy/Pandas for data processing.  
Access requests or collaboration inquiries should be directed to the corresponding author and will be evaluated through institutional review and confidentiality approval procedures.

This study was reviewed for compliance with institutional data-handling standards; no human or policyholder data were used.

\bibliography{references}
\appendix
\renewcommand{\thesection}{Appendix~\Alph{section}}
\setcounter{table}{0}
\setcounter{figure}{0}
\renewcommand{\thetable}{A\arabic{table}}
\renewcommand{\thefigure}{A\arabic{figure}}

\section{Operational Controls for Data Lineage, Integrity, and Protection}
\label{app:data_controls}

This appendix elaborates Pillar~2 (\emph{Data Lineage, Integrity, and Protection}) of the five-pillar prudential framework introduced in Section~\ref{sec:five_pillars}.  
It details implementable data-governance controls ensuring that all information exposed to large language models (LLMs) remains verifiable, auditable, and confidential throughout the lifecycle.  
These controls reflect supervisory expectations articulated by EIOPA~(2025), NAIC~(2023), and IAIS~(2024) for AI systems materially influencing underwriting, claims, or capital-model reporting~\citep{EIOPA2025,NAIC2023,IAIS2024}.  
Each control is designed to be evidence-generating—producing artifacts such as logs, lineage registries, and approval records suitable for supervisory inspection.

\subsection{Data Lineage and Provenance Logging}
\label{app:data_lineage}

Reconstructable lineage is a precondition for prudential accountability.  
All datasets exposed to LLMs should be registered in an immutable lineage store that records the full transformation chain from source treaty to model output:

\begin{enumerate}[label=(\alph*)]
    \item \textbf{Metadata capture:} each ingestion, retrieval, or augmentation logs dataset identifiers, timestamps, data owners, preprocessing steps, and retention rules.
    \item \textbf{Immutable lineage register:} entries are anchored using cryptographic hashes or equivalent write-once mechanisms so that auditors can replay every event from ingestion to output.
    \item \textbf{Bidirectional traceability:} each model output retains identifiers linking back to the treaty fragments, submissions, or regulatory texts consulted, enabling clause-level verification~\citep{Henderson2024trustnlp,Weidinger2022taxonomy}.
\end{enumerate}

This structure satisfies RAIRAB’s reproducibility requirement (Online Appendix~\ref{app:rairab}), ensuring each retrieved or generated element is attributable to a controlled, auditable source.

\subsection{Data Validation and Quality Assurance}
\label{app:data_validation}

Lineage must be complemented by rigorous data-quality assurance.  
Data entering retrieval or fine-tuning pipelines are validated at three levels:

\begin{enumerate}[label=(\alph*)]
    \item \textbf{Schema and semantic checks:} submissions, bordereaux, and exposure schedules are validated against open standards (e.g., ACORD, IFRS~17-compatible layouts) to prevent incomplete or context-free entries.
    \item \textbf{Bias and anomaly screening:} statistical profiling and rule-based filters detect geographic, peril-based, or cedent-concentration skews that could distort LLM responses.
    \item \textbf{Retrieval audits:} periodic sampling verifies that retrieved items match regulator-approved or curated treaty sets, detecting index drift or silent corruption~\citep{Balona2024actuarygpt,Gao2023survey}.
\end{enumerate}

These checks provide the empirical quality evidence expected by EIOPA~(2025) and IAIS~(2024) for AI-assisted reporting and capital-assessment systems.

\subsection{Confidentiality and Access Control}
\label{app:confidentiality}

Treaty, claims, and exposure artifacts may contain cedent-identifying or commercially sensitive information.  
Access must therefore follow least-privilege principles and generate complete audit trails:

\begin{enumerate}[label=(\alph*)]
    \item \textbf{Role-based access control (RBAC):} permissions are defined by function (actuarial, AI/ML, compliance, audit); every read/write event is logged and reviewable.
    \item \textbf{Encryption:} data are encrypted at rest and in transit (AES-256, TLS~1.3), with key management governed by enterprise cryptographic policy.
    \item \textbf{Confidential execution:} sensitive operations (fine-tuning, re-indexing) run inside trusted execution environments to prevent cross-tenant leakage.
    \item \textbf{Privacy-preserving transforms:} identifiers are masked or perturbed (e.g., differential privacy or token-level obfuscation) before exposure to LLMs~\citep{Noguer2025fair,Henderson2024trustnlp}.
\end{enumerate}

These measures are consistent with Solvency II’s data-handling obligations and RAIRAB’s conditional-access design (Online Appendix~\ref{app:rairab}).

\subsection{Prompt and Retrieval Security}
\label{app:prompt_security}

LLM deployments in multi-user or multi-client environments must be hardened against prompt injection and cross-domain leakage:

\begin{enumerate}[label=(\alph*)]
    \item \textbf{Prompt sanitization:} inputs and retrieved text are scanned for instruction-injection patterns and disallowed directives.
    \item \textbf{Index partitioning:} retrieval stores are segregated by client, business line, or jurisdiction; queries are routed only to authorized partitions.
    \item \textbf{Leakage and inversion monitoring:} abnormal token distributions or repeated near-duplicate queries trigger alerts for potential exfiltration.
    \item \textbf{Credential governance:} access tokens are rotated and every prompt–response pair is tagged with user identity, timestamp, and retrieval context.
\end{enumerate}

These protections allow RAIRAB evaluations to be replicated under supervisory review without exposing sensitive cedent data.

\subsection{Monitoring, Incident Response, and Reporting}
\label{app:monitoring}

Continuous monitoring integrates these data controls into the broader prudential-governance loop:

\begin{enumerate}[label=(\alph*)]
    \item \textbf{Lineage dashboards:} governance teams monitor lineage completeness, validation pass rates, and exception volumes.
    \item \textbf{Automated escalation:} failed validations, unauthorized retrievals, or abnormal retrieval drift generate incident reports routed to data owners and model-risk functions.
    \item \textbf{Audit-ready exports:} periodic snapshots of lineage logs, access trails, and retrieval proofs are packaged for supervisory examination under EIOPA, IAIS, or NAIC standards~\citep{EIOPA2025,NAIC2023,IAIS2024}.
\end{enumerate}

These processes demonstrate \emph{continuous assurance}—a key expectation in recent BIS~(2024) and FSB~(2025) governance principles.

\subsection{Relation to the Five-Pillar Framework and RAIRAB}
\label{app:relation}

Controls defined in Sections~\ref{app:data_lineage}–\ref{app:monitoring} operationalize Pillar~2 and reinforce the remaining four pillars:  
Pillar~1 assigns ownership and accountability;  
Pillar~3 consumes lineage artifacts as assurance evidence;  
Pillar~4 relies on monitored drift recovery; and  
Pillar~5 ensures lawful, fair, cross-jurisdictional handling of reinsurance data.  
The de-identified corpus used for RAIRAB (Section~\ref{sec:RAIRAB}) was prepared under these safeguards.

\begin{figure}[htbp!]
\centering
\begin{tikzpicture}[
    scale=0.78, transform shape,
    node distance=1.8cm and 1.4cm,
    >=latex, font=\footnotesize,
    proc/.style={rectangle, rounded corners=4pt, draw=black, very thick,
                 fill=gray!6, minimum width=3.3cm, minimum height=1.05cm,
                 align=center},
    store/.style={rectangle, rounded corners=4pt, draw=black,
                  fill=gray!15, minimum width=3.25cm, minimum height=0.9cm,
                  align=center},
    line/.style={->, very thick}
]

\node[store] (sources) {Treaty / Claims /\\ Exposure Sources};
\node[proc, below=of sources, xshift=-2.7cm] (lineage) {Provenance \&\\ Lineage Capture};
\node[proc, below=of sources, xshift=2.7cm] (validation) {Data Quality \&\\ Semantic Checks};
\node[proc, below=2.0cm of lineage] (access) {RBAC \&\\ Policy Enforcement};
\node[proc, below=2.0cm of validation] (llm) {Governance-Aware\\ LLM / RAG Service};
\node[store, below=1.9cm of access] (logging) {Audit \& Usage Logs\\ (Clause-Level)};
\node[store, below=1.9cm of llm] (reporting) {Supervisory /\\ Internal Reporting};

\draw[line] (sources.south) -- (lineage.north);
\draw[line] (sources.south) -- (validation.north);
\draw[line] (lineage.south) -- (access.north);
\draw[line] (validation.south) -- (llm.north);
\draw[line] (access.east) -- (llm.west);
\draw[line] (access.south) -- (logging.north);
\draw[line] (llm.south) -- (reporting.north);
\draw[line, dashed] (logging.east) -- (reporting.west);

\node[align=left, anchor=west] at ([xshift=0.25cm]llm.east) {%
  \begin{tabular}{@{}l@{}}
    \textit{Captured artifacts:}\\
    -- prompt / response logs\\
    -- retrieval IDs\\
    -- human overrides
  \end{tabular}
};

\node[align=left, anchor=west] at ([xshift=0.25cm]reporting.east) {%
  \begin{tabular}{@{}l@{}}
    \textit{Outputs:}\\
    -- Solvency II / EIOPA reports\\
    -- NAIC / IAIS templates\\
    -- internal audit exports
  \end{tabular}
};

\end{tikzpicture}
\caption{Operational data-control flow for Pillar 2 (Data Lineage, Integrity, and Protection).  
Source artifacts are captured with provenance metadata, validated for quality, and accessed only through governed LLM/RAG services.  
All interactions are logged and exportable for supervisory review.}
\label{fig:data_control_flow}
\end{figure}
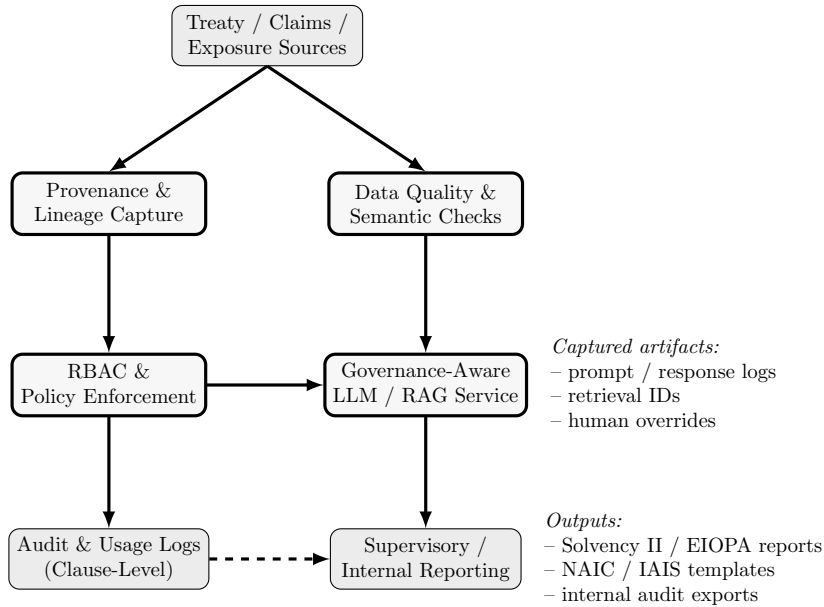

\section{Empirical Evaluation Design: The Reinsurance AI Reliability and Assurance Benchmark (RAIRAB)}
\label{app:rairab}

This appendix documents the empirical design, scoring methodology, and access conditions of the \emph{Reinsurance AI Reliability and Assurance Benchmark} (RAIRAB) introduced in Section~\ref{sec:RAIRAB}.  
RAIRAB operationalizes the five-pillar prudential framework as a reproducible evaluation suite measuring whether governance-embedded large language models (LLMs) meet supervisory expectations for traceability, interpretive stability, and documentation in solvency-relevant workflows.  
The benchmark aligns with assurance guidance from EIOPA~(2025), IAIS~(2024), NAIC~(2023), and BIS/FSB~(2024–2025), and was developed on a de-identified, regulator-conformant corpus to ensure confidentiality and audit integrity~\citep{EIOPA2025,NAIC2023,IAIS2024,BIS2024AI,FSB2025AI}.

\subsection{Scope, Tasks, and Corpus Construction}
\label{app:rairab_scope}

RAIRAB covers six prudentially material task families whose outputs influence pricing, claims, or solvency reporting:
\begin{enumerate}[label=(\alph*)]
    \item treaty clause interpretation,
    \item submission summarization,
    \item endorsement and wording-drift detection,
    \item claims validation,
    \item exposure extraction and accumulation analysis,
    \item compliance and capital-documentation generation.
\end{enumerate}

The corpus integrates open exposure templates (OED, CEDE, ACORD) with synthetically generated treaty and claims artifacts, ensuring realism without breaching confidentiality.  
Structured components—attachments, limits, reinstatements, and peril fields—derive from the \emph{Open Exposure Data (OED)} schema maintained by Oasis~\citep{OED2024}, Verisk’s \emph{Catastrophe Exposure Data Exchange (CEDE)}~\citep{Verisk2024CEDE}, and the \emph{ACORD Reinsurance Data Standards (2024)}~\citep{ACORD2024}.  
Textual artifacts—treaty clauses, endorsements, and loss narratives—were generated via LLMs conditioned on these schemas and public exemplars from EIOPA~(2025), NAIC~(2023), and IAIS~(2024).  
All identifiers were masked, and metadata (prompt template, random seed, schema version) were logged to satisfy Solvency II Article 41 lineage standards.

Approximately half of the corpus is synthetic; the remainder comprises template-based reconstructions derived from open supervisory exemplars rather than proprietary contracts.  
Each record is versioned, cryptographically hashed, and registered in a lineage ledger to ensure full audit traceability.  
This hybrid design balances representativeness, confidentiality, and reproducibility—key principles for prudential validation research.

\subsection{Model Families and Controlled Configurations}
\label{app:rairab_models}

Six model families—GPT-4/5, Claude 3/3.5, Gemini 1.5 Pro, LLaMA 3, Mistral 8$\times$22B, and DeepSeek V2—were evaluated under three governance regimes:
\begin{enumerate}[label=(\roman*)]
    \item \emph{Zero-shot prompting:} task description only;  
    \item \emph{Few-shot prompting:} three to five fixed in-domain exemplars per task;  
    \item \emph{RAG + logging + HITL:} retrieval-augmented generation with clause-level retrieval IDs, structured logging, and human-in-the-loop (HITL) review checkpoints.
\end{enumerate}

All experiments used a temperature setting of $T=0.2$, a 16k-token context window, and isolated environments with no external API access.  
Governance artifacts—including retrieval IDs, version tags, and reviewer approvals—were automatically logged for audit replication.  
These settings correspond to the governance distinctions formalized in Section~\ref{subsec:rairab_protocol} and reflect the control expectations outlined in the BIS~(2024) \emph{Principles for Model Risk Management}~\citep{BIS2024AI}.

\subsection{Metrics and Linkage to the Five-Pillar Framework}
\label{app:rairab_metrics}

RAIRAB reports five prudential metrics, each linked to a corresponding pillar in Section~\ref{sec:five_pillars}:

\begin{itemize}
    \item \textbf{Grounding Accuracy (GA)} — Pillar 2: degree to which model outputs are supported by controlled sources and lineage references.
    \item \textbf{Hallucination Rate (HR)} — Pillar 3: frequency of unsupported or extraneous content requiring audit remediation~\citep{Ji2023hallucination,OpenAI2025Hallucinate}.
    \item \textbf{Interpretive Drift (ID)} — Pillar 4: variability across identical prompts, capturing semantic stability and resilience.
    \item \textbf{Transparency Index (TI)} — Pillar 3: visibility of citations, rationale, and log completeness for replayable supervision.
    \item \textbf{Compliance Alignment (CA)} — Pillar 1: evidence of governance artifacts (segregation, approvals, and change-control logs).
\end{itemize}

Each model–task–configuration combination was executed three times; 95\% confidence intervals were estimated through non-parametric bootstrapping (1,000 resamples), and all metrics were normalized to $[0,1]$.  
Metric rationales follow the prudential-measurement tradition of \citet{CumminsWeiss2014} and align with assurance frameworks proposed by \citet{EIOPA2025} and \citet{IAIS2024}.

\subsection{Human Adjudication and Baselines}
\label{app:rairab_human}

A panel of underwriters, actuaries, and compliance specialists from the EU, UK, and US reviewed approximately 10\% of outputs, blinded to model identity.  
Disagreements were adjudicated by a third reviewer, producing inter-rater reliability $\kappa = 0.87$.  
All reviewers were domain practitioners trained using a standardized rubric covering grounding, regulatory language fidelity, and interpretive consistency.  
Calibration involved ten gold-labeled examples derived from open templates prior to scoring.  
This adjudication protocol adheres to actuarial validation practices and the transparency standards promoted by EIOPA~(2025) and IAIS~(2024)~\citep{EIOPA2025,IAIS2024,BIS2024AI}.

A deterministic \emph{document-intelligence baseline}—built from rule-based extraction, regular expressions, and clause templates—served as a benchmark for current underwriting-automation systems.  
While precise on structured data, it produced no governance artifacts, explaining its lower Transparency (TI) and Compliance Alignment (CA) scores.

\subsection{Reproducibility and Access Model}
\label{app:rairab_access}

Because the corpus includes de-identified but prudentially sensitive materials, it cannot be fully public.  
However, benchmark templates, scoring rubrics, and orchestration scripts can be shared under confidentiality and jurisdictional-compliance agreements upon reasonable request.  
All datasets are versioned with SHA-256 lineage hashes and stored in encrypted repositories to support audit re-execution.  
This conditional-access approach reflects EIOPA’s \emph{proportional-disclosure} principle and IAIS guidance on \emph{supervisory transparency without public data release}~\citep{EIOPA2025,IAIS2024}.  

Together, these procedures ensure that RAIRAB satisfies both research reproducibility and prudential data-protection standards, enabling evidence-based supervision while preserving confidentiality.

\section{Supervisory Frameworks Referenced in the Paper}
\label{app:supervisory_table}

This appendix consolidates the prudential and supervisory frameworks cited throughout Sections~\ref{sec:five_pillars}–\ref{sec:policy}.  
It provides a single reference point for reviewers and practitioners, illustrating the global convergence of principles that underpin the proposed five-pillar governance architecture, the operational controls in Online Appendix~\ref{app:data_controls}, and the RAIRAB benchmark introduced in Section~\ref{sec:RAIRAB}.  
The table highlights how these frameworks collectively align around lifecycle governance, accountability, transparency, and data integrity for AI systems influencing solvency-relevant decisions.

\begin{table}[htbp!]
\centering
\footnotesize
\caption{Principal supervisory frameworks relevant to LLM-enabled reinsurance governance (indicative, not exhaustive).}
\renewcommand{\arraystretch}{1.2}
\setlength{\tabcolsep}{5pt}
\begin{tabular}{p{4.5cm}p{3.5cm}p{7.3cm}}
\toprule
\textbf{Framework} & \textbf{Jurisdiction / Year} & \textbf{Key implications for AI assurance and governance} \\
\midrule
\emph{EIOPA Opinion on AI Governance and Risk Management} & EU / 2025 &
Establishes lifecycle oversight, verifiable data lineage, and human-in-the-loop (HITL) controls for AI systems affecting underwriting, pricing, or capital modeling~\citep{EIOPA2025}. \\

\emph{NAIC Model Bulletin on the Use of AI Systems by Insurers} & US / 2023 &
Requires enterprise-level AI-governance programs, bias and fairness testing, and documentation of validation and accountability structures~\citep{NAIC2023}. \\

\emph{IAIS Application Paper on the Supervision of AI and Machine Learning} & Global / 2024 &
Defines proportionality, explainability, and human accountability as minimum supervisory expectations for AI in insurance markets~\citep{IAIS2024}. \\

\emph{SR\,11-7 Model Risk Management Guidance} & US / 2011 &
Sets standards for model inventory, independent validation, and change control—principles directly extendable to LLM-based decision frameworks. \\

\emph{Solvency II Directive, Pillar 2 (Governance and Risk Management)} & EU / 2016 &
Requires board-level accountability, documentation, and internal-model validation, forming the prudential analogue for AI lifecycle control. \\

\emph{BIS, FSB, and NGFS Reports on AI Assurance and Model Governance} & Global / 2024–2025 &
Emphasize transparency, operational resilience, continuous validation, and systemic-risk mitigation for AI models used in regulated financial institutions~\citep{BIS2024AI,FSB2025AI,NGFS2024scenarios}. \\
\bottomrule
\end{tabular}
\end{table}

\end{document}